\documentclass{article}

     \usepackage[final]{ml4mols_2020}

\usepackage[utf8]{inputenc} 
\usepackage[T1]{fontenc}    
\usepackage{hyperref}       
\usepackage{url}            
\usepackage{booktabs}       
\usepackage{amsfonts}       
\usepackage{nicefrac}       
\usepackage{microtype}      
\usepackage{amsmath}
\usepackage{graphicx}

\title{Auto-Encoding Molecular Conformations
}

\author{%
  Robin Winter\\
  Machine Learning Research\\
  Bayer AG\\
  Berlin\\
  \texttt{robin.winter[at]bayer.com} \\
  \And
  Frank No\'{e} \\
  Mathematics and Computer Science\\
  Freie Universit\"{a}t Berlin\\
  Berlin
  \And
  Djork-Arn\'{e} Clevert\\
  Machine Learning Research\\
  Bayer AG\\
  Berlin\\
  \texttt{djork-arne.clevert[at]bayer.com}
}

\begin{document}

\maketitle

\begin{abstract}
  In this work we introduce an Autoencoder for molecular conformations. Our proposed model converts the discrete spatial arrangements of atoms in a given molecular graph (conformation) into and from a continuous fixed-sized latent representation. We demonstrate that in this latent representation, similar conformations cluster together while distinct conformations split apart. Moreover, by training a probabilistic model on a large dataset of molecular conformations, we demonstrate how our model can be used to generate diverse sets of energetically favorable conformations for a given molecule. Finally, we show that the continuous representation allows us to utilize optimization methods to find molecules that have conformations with favourable spatial properties.
\end{abstract}

\section{Introduction}
Representing chemical matter in a meaningful and expressive way plays a crucial role when it comes to computer-aided modeling in the field of chemistry \citep{todeschini2009molecular}. Recently, substantial progress has been made in many molecule-related tasks, such as bio- and physicochemical property prediction  \citep{montanari2020}, inverse design of desirable molecules \citep{gomez2018automatic, winter2019efficient, Le2020}, 
retrosynthetic analysis and synthesis planning \citep{segler2018}. Most of these advancements can be attributed to the use of deep neural networks that enable representation learning of chemical matter. While traditional methods are mostly based on features, generated by rule-based algorithms extracting structural information (e.g. extended-connectivity fingerprints), these novel methods are directly trained on a discrete but more comprehensive representation of molecules. In particular, Graph Neural Networks utilizing the molecular graph with atoms as nodes and bonds as edges \citep{duvenaud2015convolutional} or Recurrent Neural Networks utilizing the one-dimensional line notation of molecules, namely SMILES \citep{segler2018generating}. Still, the underlying molecular representation of the aforementioned methods are limited in the sense that they do not reflect the spatial arrangement of the atoms in the molecule. However, many interesting molecular properties, such as its ability to fit inside a protein binding pocket, inducing a pharmacological effect, are driven by its possible (energetically stable) spatial arrangements (conformations). Recently, work has been done to apply neural networks directly on specific conformations of molecules to predict properties such as the equilibrium energy or the \emph{HOMO-LUMO} gap \citep{schutt2018schnet}. Moreover, models have been proposed to generate molecular conformations for a given molecular graph \citep{mansimov2019molecular, simm2019generative} or chemical formula \citep{hoffmann2019generating}. \\
In this work we propose a novel model that converts a molecular conformation to and from a fixed-sized latent representation (conformation embedding), independent of the number of atoms and bonds of a molecule. Moreover, training the model in a probabilistic setting, we show that we can model the conformational distributions of molecules. We demonstrate that sampling from this model results into a diverse set of energetically reasonable conformers. Finally, combining the conformation embedding with a continuous molecular structure embedding, we demonstrate how molecules can be optimized with respect to spatial properties.

\begin{figure}
\label{network_arc}
\includegraphics[scale=0.41]{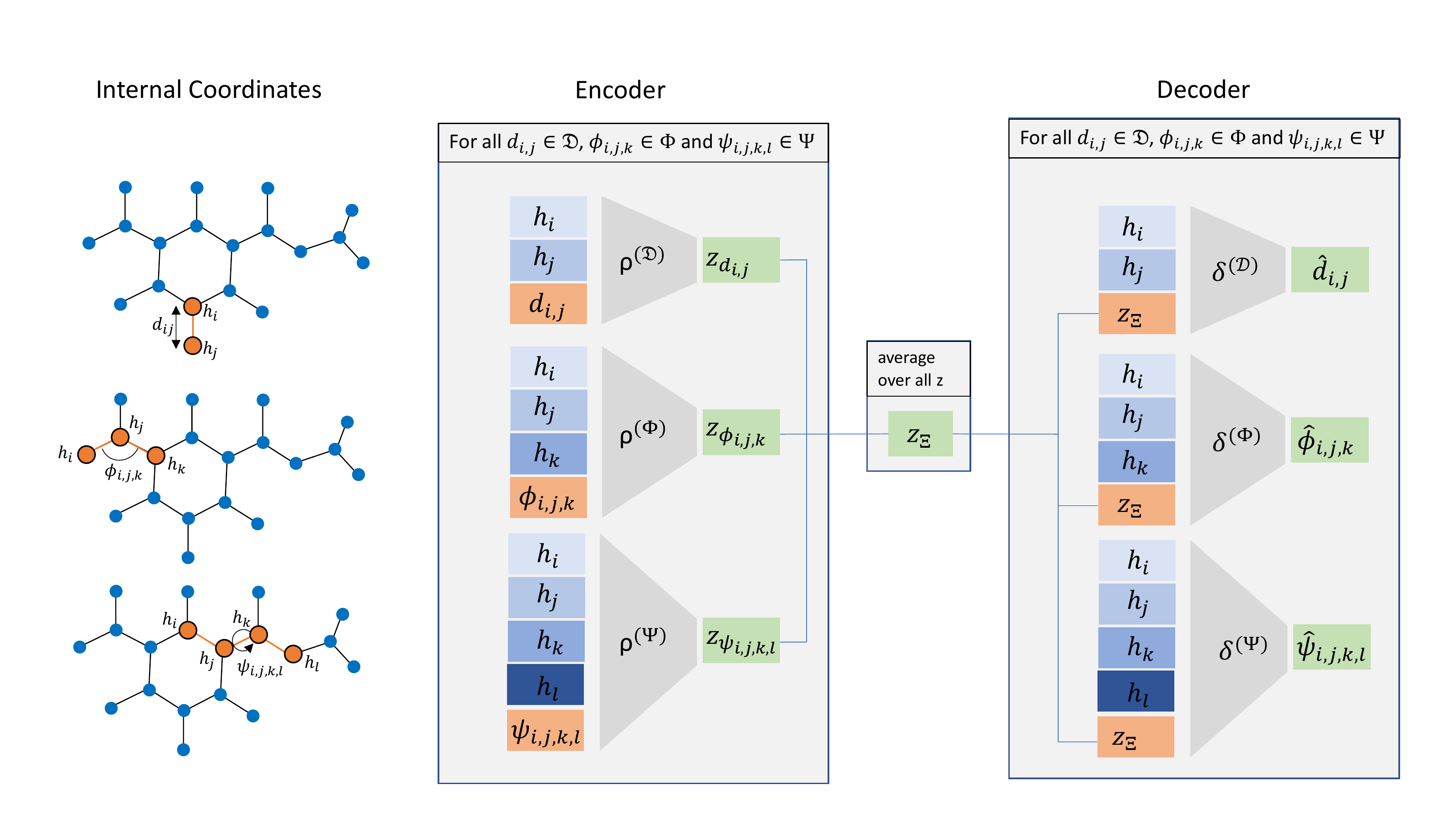}
\caption{Model architecture of the conformation encoder (middle) and decoder (right). The encoding functions $\rho^{(\mathcal{D})}$, $\rho^{(\Phi)}$ and $\rho^{(\Psi)}$ encode their respective internal coordinates into a latent representation. The decoding functions $\delta^{(\mathcal{D})}$, $\delta^{(\Phi)}$ and $\delta^{(\Psi)}$ are conditioned on the averaged latent representations (conformer embedding) to reconstruct their respective internal coordinates, given a set of node embeddings $h_i\in\mathcal{H}$. On the left hand side, the definition of the internal coordinates, bond length $d_{i,j}\in\mathcal{D}$, bond angle $\phi_{i,j,k}\in\Phi$  and dihedral angle $\psi_{i,j,k,l}\in\Psi$, is visualized.}
\end{figure}

\section{Methods}
\subsection{Representing Molecular Conformations}
The three-dimensional arrangement of atoms of a molecule can be represented in many different ways. Annotating each atom with a Cartesian coordinate is probably the most straight-forward way, however, does not reflect a molecules invariance to rigid translations and rotations. In this work we utilize the \emph{internal coordinate representation}, also known as \emph{Z-matrix}. In this notation, a molecules spatial arrangement (conformation) $\Xi$ is defined by the set of distances $\mathcal{D}=\{d_1,\ldots,d_{N_\mathcal{D}}\}$ between bonded atoms (bond length), the angles $\Phi=\{\phi_1,\ldots,\phi_{N_\Phi}\}$ of three connected atoms (bond angles) and the torsion angles (dihedral angles) $\Psi=\{\psi_1,\ldots,\psi_{N_\Psi}\}$ of three consecutive bonds (see Figure \ref{network_arc}). This representation is invariant to rotations and rigid translations and can always be transformed to and from Cartesian coordinates.
\subsection{Conformation Autoencoder}
\label{conf_ae}
The overall goal of the proposed model is to find functions $f_\Theta$ and $g_\Theta$ that map a conformation $\Xi_\mathcal{G}$ of a molecule $\mathcal{G}$ to and from a fixed-sized latent representation $z_\Xi\in\mathbb{R}^{F_z}$, respectively. This introduces two major challenges. Firstly, the model has to map molecules with a different number of atoms and bonds to the same fixed-sized space. Secondly, $f_\Theta$ has to invariant with respect to the ordering of atoms in the input. Our proposed model consist of a conformation-independent and a conformation-dependent part. The conformation-independent part comprises a Graph Neural Network utilizing the molecular graph to extract node-level features for a given molecule. The conformation-dependent part utilizes these extracted node-level features either to encode the internal coordinates of a specific molecular conformation into a latent representation (conformation embedding) or to reconstruct a conformation by predicting the internal coordinates of sets of connected atoms, given their respective node features and a conformation embedding. The whole model is trained on minimizing the reconstruction error of the internal coordinates $\Xi$ for a given molecule:
\begin{equation}
\label{cost}
    \mathcal{C}_\Xi=\frac{1}{N_\mathcal{D}}\sum_{d\in\mathcal{D}}\Vert d, \hat{d}\Vert_2^2 +
    \frac{1}{N_\Phi}\sum_{\phi\in\Phi}\Vert \phi, \hat{\phi}\Vert_2^2 +
    \frac{1}{N_\Psi}\sum_{\psi\in\Psi}\text{min}((\Vert \psi, \hat{\psi}\Vert_2^2, 2\pi-\Vert \psi, \hat{\psi}\Vert_2^2),
\end{equation}
where the last term accounts for the periodicity of the dihedral angle. 
Our proposed model can easily be extended to a probabilistic generative model by employing the ideas from \citet{kingma2013auto}, effectively defining the model as a variational auto encoder.
\subsubsection{Molecular Graph Encoder}
\label{graph encoder}
In this work we define the conformation-independent molecular graph as an undirected Graph $\mathcal{G}=(\mathcal{V},\mathcal{E})$, with vertices or nodes $v_i\in \mathcal{V}$ and edges $e_{ij}=(v_i,v_j)\in \mathcal{E}$ connecting $v_i$ and $v_j$. Where nodes $v_i \in\mathbb{R}^{F_v}$ represent atoms and encode atom features such as element type and charge. Edges $e_{ij}\in\mathbb{R}^{F_e}$ represent bonds between atoms and encode the bond type  (i.e. single, double, triple or aromatic bond).\\
We utilize a Graph Neural Network (GNN) to extract a node-level representation of a molecular graph. Given a molecular graph with initial node and edge features defined by the atoms and bonds of the molecule, a GNN iteratively updates node embeddings by aggregating localized information of connected nodes respectively.

In particular, we use the Graph Attention Network (GAT) \citep{velivckovic2017graph} framework which updates the node embeddings $\mathbf{h}_i$ by the following rule:
\begin{equation}
    \mathbf{h}^{'}_i = \alpha_{i,j}\mathbf{\Theta} \mathbf{h}_i + \sum_{j \in \mathcal{N}(i)} \alpha_{i,j}\mathbf{\Theta} \mathbf{h}_j,
\end{equation}
with neighbouring node indices $\mathcal{N}(i)=\{j\in(0,\ldots,N)\vert\forall j ((v_i, v_j) \in \mathcal{E})\}$. The attention coefficients $\alpha_{i,j}$ are computed as
\begin{equation}
\alpha_{i,j} = \frac{\exp(\sigma(\textbf{a}^T[\mathbf{\Theta}\mathbf{h}_i\Vert\mathbf{\Theta}\mathbf{h}_j]))}{\sum_{k\in \mathcal{N}(i)\cup\{i\}}\exp(\sigma(\mathbf{a}^T[\mathbf{\Theta}\mathbf{h}_i\Vert\mathbf{\Theta}\mathbf{h}_k]))},
\end{equation}
where $\cdot^T$ is the  transposition and $\Vert$ the concatenation operation. The attention function $a$ is defined as a single-layer feedforward neural network. \\
To incorporate edge attributes $\mathbf{e}_{i,j}$ (bond-type information) in the model we also utilize the so-called edge-conditioned graph convolution (EConv) layer \citep{simon}, defined by the following update rule:
\begin{equation}
\mathbf{h}^{'}_i = \mathbf{\Theta} \mathbf{h}_i + \sum_{j \in \mathcal{N}(i)} \mathbf{h}_j \cdot \text{f}_{\mathbf{\Theta}}(\mathbf{e}_{i,j}),
\end{equation}
where $\text{f}_{\mathbf{\Theta}}$ denotes a multi layer perceptron.

In particular, we utilized edge-conditioned graph convolution (EConv) \citep{simon} and Graph Attention Network (GAT) \citep{velivckovic2017graph} layers.
To aggregate information about higher-order neighbours, we combine one EConv (to encode edge information) with multiple consecutive GAT layers:
\begin{equation}
\mathbf{h}_i^{l} = \text{GAT}^{l-1}\circ\ldots\circ\text{GAT}^{1}\circ\text{EConv}(\mathbf{h}_i^{0}).
\end{equation}
where $\mathbf{h}_i^{0} = v_i \in\mathbb{R}^{F_v}$ are the atom features of the input molecular graph.
\subsubsection{Conformation Encoder}
To this end, we define the molecular conformation representation learning task as a learning task on \emph{sets}. In particular, our proposed model learns to extract a latent representation $z_\Xi$ of a set of internal coordinates $\Xi$ for a given molecule:
\begin{equation}
    z_\Xi = f_{\mathbf{\Theta}}(\mathcal{H},\Xi) = f_{\mathbf{\Theta}}(\mathcal{H}, \mathcal{D}, \Phi, \Psi),
\end{equation}
with the permutation invariant function $f_{\mathbf{\Theta}}$, parameterized by a neural network and conditioned on the node embeddings $\mathcal{H}=\{h_1,\ldots h_N\}$, extracted by the molecular graph encoder defined in the previous section. Following \cite{zaheer2017deep}, we define the permutation invariant function $f_{\mathbf{\Theta}}$ as
\begin{equation}
\label{enocder_eq}
\begin{split}
z_\Xi &= f_{\mathbf{\Theta}}(\mathcal{H},\Xi) = \sigma\left(\sum_{\xi\in\Xi}\rho(\mathcal{H},\xi)\right)=\frac{1}{N_\Xi}\sum_{\xi\in\Xi} \rho_\Theta(\mathcal{H},\xi) \\
&= \frac{1}{N_\mathcal{D} + N_\Phi + N_\Psi}\left(\sum_{d\in\mathcal{D}}\rho^{(\mathcal{D})}_\Theta(\mathcal{H}, d) + \sum_{\phi\in\Phi}\rho^{(\Phi)}_\Theta(\mathcal{H}, \phi) + \sum_{\psi\in\Psi}\rho^{(\Psi)}_\Theta(\mathcal{H}, \psi)\right).
\end{split}
\end{equation}
The functions $\rho^{(\mathcal{D})}_\Theta$, $\rho^{(\Phi)}
_\Theta$ and $\rho^{(\Psi)}_\Theta$ are defined as feed-forward neural networks that take bond lengths, bond angles and dihedral angles as input respectively. Additionally, to put an internal coordinate into context of the graph, $\rho_\Theta$ is conditioned on corresponding node embeddings $\mathcal{H}$ as well. This means, if $\rho^{(\Phi)}_\Theta$ encodes the angle $\phi_{i,j,k}$ between atoms $v_i$ and $v_k$ connected by atom $v_j$, function $\rho^{(\Phi)}_\Theta$ takes also node embeddings $h_i$, $h_j$ and $h_k$ as argument (see Figure \ref{network_arc}). Most importantly, equation \eqref{enocder_eq} is invariant to the order of internal coordinates and node indices and the dimensionality of the resulting $z_\Xi$ is invariant of the size of the input molecule.  
\subsubsection{Conformation Decoder}
To reconstruct a molecular conformation back from its latent representation, we train three additional neural networks $\delta^{(\mathcal{D})}_\Theta$, $\delta^{(\Phi)}
_\Theta$ and $\delta^{(\Psi)}_\Theta$ for each type of internal coordinate respectively (see Figure \ref{network_arc}). Each neural network is conditioned on the conformation embedding and takes the node embeddings of the corresponding internal coordinate as input. For example, to predict the bond length $d_{i,j}$ between atoms $v_i$ and $v_j$, the network takes $h_i$, $h_j$ and $z_\Xi$ as input.

\begin{figure}
    \centering
    \includegraphics[trim=2cm 2cm 2cm 4cm, scale=0.42]{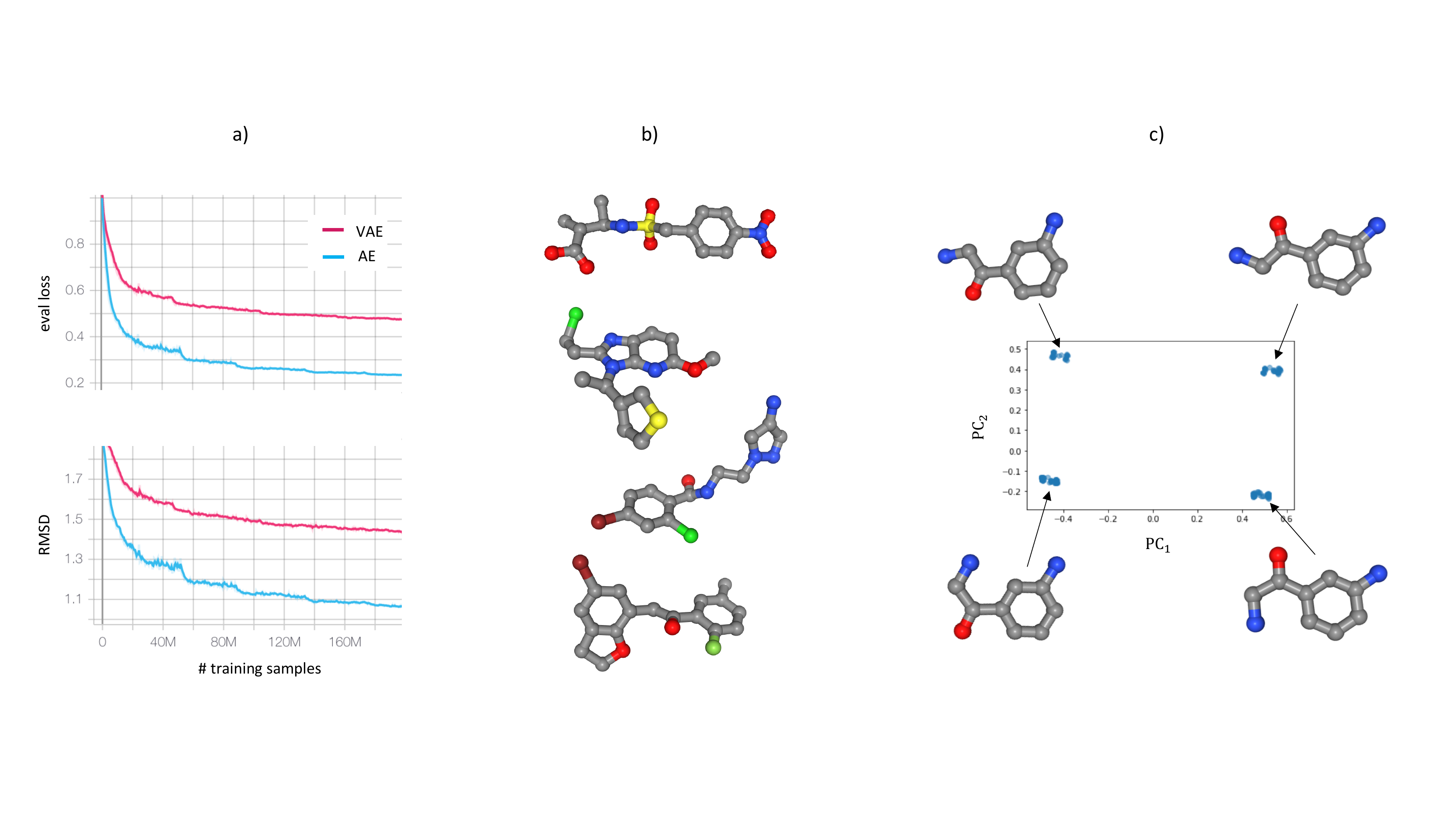}
    \caption{a) Learning curves for the autoencoder (AE) and variational autoencoder (VAE) during training, with evaluation loss as defined in \eqref{cost} and root means squared deviation (RMSD) between predicted and input conformations on a holdout set. b) Four example conformations generated by our proposed model. c) First two principle components of the latent representation (conformation embedding) of 200 conformations with a corresponding representative conformation for each cluster.}
    \label{results}
\end{figure}

\section{Results and Discussion}
We trained our model on the public PubChem3D dataset \citep{bolton2011pubchem3d}, which comprises molecules (organic, up to 50 heavy atoms) with multiple conformations generated by the forcefield software OMEGA \citep{hawkins2010conformer}. Upon convergence, our model is able to predict internal coordinates for a given molecule that result into conformations that are similar (with respect to the RMSD) to the input conformations (see Figure \ref{results}). To quantitatively analyze how energetically reasonable the reconstructed conformations are, we calculated their internal energy with the MMFF94 forcefield \citep{halgren1996merck} as implemented in the Python package RDKit \citep{landrum2006rdkit}. The median energetic difference between the input and reconstructed conformation is approximately $80\, \nicefrac{\text{kcal}}{\text{mol}}$, which corresponds to small deviations from local minimas, without e.g. clashes of atoms (see example molecules in Figure \ref{results}).
Moreover, since the model does not only reconstruct any possible conformation for a molecule but is trained on reconstructing a specific input conformation, differences between these conformations have to be encoded in the latent representation. On the right side of Figure \ref{results}, we show this for a simple example of a small molecule in four different conformations.\\
As described in Section \ref{conf_ae}, we can easily extend the proposed model to a variational autoencoder, which can be used to sample conformations from the learned distribution. A major challenge in conformation generation is to efficiently sample diverse conformers. Therefore, we analyzed the average interconformer RMSD (icRMSD) for a set of 200 sampled conformers per molecule for the holdout set. Comparing the icRMSD of our proposed model with a state-of-the-art conformation generation algorithm ETKDG \citep{riniker2015better} as implemented in RDKit, we see a similar performance, with our model having a slightly higher average icRMSD of $0.07\,$\AA. \\
Since the proposed model gives means to directly infer conformations for a given molecule, it is possible to optimize molecules in the continuous conformation embedding with respect to spatial properties. When combined with a latent representation of the molecular structure \citep{winter2019learning}, optimization of molecules can even be performed with respect to both the molecular graph and its conformation. As a proof of principle, we optimized molecules with respect to a combination of the conformation-independent \emph{quantitative estimate of drug-likeness} (QED) score (values between 0 and 1) \citep{bickerton2012quantifying} and the conformation-dependent property \emph{asphericity} \citep{todeschini2009molecular} (values between 0 and 1), which quantifies a molecules deviation from a spherical shape. We utilized the genetic \emph{Particle Swarm Optimization} algorithm \citep{kennedy1995particle}, to optimize both latent representations at the same time. Starting from the already drug-like molecule aspirin with a combined score of $0.76$, we could already find after 50 iterations molecules with a score of $1.82$. In general this method could also be used to optimize molecules for other interesting spatial properties, such as fitting pharmacophores or the shape of known bio active molecule.

\bibliography{bib}
\bibliographystyle{achemso}
\end{document}